\documentclass[letterpaper, 10 pt, conference]{ieeeconf}  

\IEEEoverridecommandlockouts                              

\usepackage{graphics} 
\usepackage{graphicx}
\usepackage{epsfig} 
\usepackage{amsmath} 
\usepackage{amssymb}  
 \usepackage{algorithm}
\PassOptionsToPackage{noend}{algpseudocode}
\usepackage{algpseudocode}
\algnewcommand\And{\textbf{ and }}
\algnewcommand\Or{\textbf{ or }}
\usepackage[font=small]{caption}
\usepackage{subcaption}
\usepackage{color}
\usepackage{url}
\usepackage{courier}

\title{\LARGE \bf
Realizing the Aerial Robotic Worker for Inspection Operations
}

\author{Kostas Alexis
\thanks{The author is with the Autonomous Robots Lab, University of Nevada, Reno,
	1664 N. Virginia Street, Reno, NV 89557, USA
	{\tt\small kalexis@unr.edu}}%
}

\begin{document}

\maketitle
\thispagestyle{empty}
\pagestyle{empty}

\begin{abstract}

This report overviews a set of recent contributions in the field of path planning that were developed to enable the realization of the autonomous aerial robotic worker for inspection operations. The specific algorithmic contributions address several fundamental challenges of robotic inspection and exploration, and specifically those of optimal coverage planning given an a priori known model of the structure to be inspected, full coverage, optimized and fast inspection path planning, as well as efficient exploration of completely unknown environments and structures. All of the developed path planners support both holonomic and nonholonomic systems, and respect the on--board sensor model and constraints. An overview of the achieved results, followed by an integrating architecture in order to enable fully autonomous and highly--efficient infrastructure inspection in both known and unknown environments. 

\end{abstract}

\section{INTRODUCTION}\label{sec:intro}

The vision of automated infrastructure monitoring and damage detection corresponds to the motivation of a body of work conducted by our research team and aims to develop the ``Aerial Robotic Worker'' (ARW) as a class of systems that --among others-- are able to autonomously conduct structural inspection operations. Infrastructure is the foundation that connects our resources, energy flows, communities, and people, driving our economy and improving our quality of life. For our economies to be sustainable, a first class infrastructure system, in terms of quality, distribution, long--term operation, systematic inspection and maintenance is required. Within recent years, the scientific breakthroughs and technological developments in the area of civilian mobile robotics have progressively brought robotics closer to real--life challenging applications, and pioneering use cases have already shown a very promising potential. Indicative scenarios include those of monitoring of bridges, solar farms, power generation and distribution facilities, geothermal facilities, oil \& gas industry infrastructure, water dams and more. As indicated in several reports such as the one provided by the American Society of Civil Engineers (ASCE)~\cite{nsf_asce_card2013}, our infrastructure is under the urgent need for a more systematic approach for the relevant inspection operations due to its degraded condition. Characteristically, the US dams are graded with $D$ by ASCE and among the $84000$ dams, $14000$ are characterized as high--hazard, while their average age is $52$ years. The $607380$ U.S. bridges are graded with $C+$, have an average age of $42$ years and $1$ out of $9$ of them are rated as structurally deficient.

%
\begin{figure}[h!]
\centering
  \includegraphics[width=0.99\columnwidth]{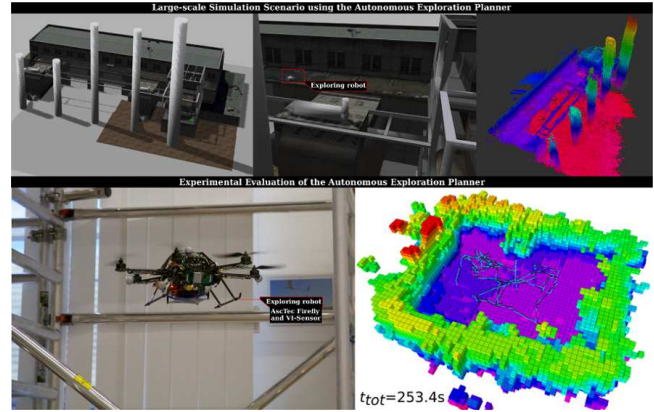}
\caption{The concept of aerial robotic workers aims to realize a class of aerial robotic systems capable of autonomous inspection of infrastructure and other critical facilities. In the images depicted, instances from a large--scale simulation--based exploration scenario as well as photo and data from an experimental study on autonomous exploration are depicted. }
\label{fig:intro}
\end{figure}
%

But to automate the inspection process, an aerial robot should be something much more than a position--controlled camera in the sky. To develop the aerial robotic worker for inspection operations, significant challenges in perception and path planning have to be addressed. Within the framework of this work, we provide an overview of a set of algorithms that were recently proposed by our team and break new ground on how an aerial robot handles efficient exploration of unknown environments, volumetric mapping and full coverage, high fidelity structural 3D reconstruction and as a last step, possibly required contact--based inspection. The proposed algorithms specifically address the problems of a) optimal full coverage of a structure for which a known geometric model exists, b) fast, optimized (but suboptimal) full coverage inspection of structures with known geometric model, c) uniform coverage of such structures, d) autonomous exploration and inspection of completely unknown environments and the structures in them, and finally e) contact--based inspection of selected points of interest on the $3\textrm{D}$ structure. Each algorithm has its advantages and disadvantages, while an overall architecture is proposed such that their combination can lead to the fully autonomous execution of inspection tasks in either known or unknown environments with selective levels of computational cost and inspection precision. It is noted that, all the proposed methods have been experimentally verified. Figure~\ref{fig:intro} presents a subset of the relevant results. Within this work, we present highlights of the previously derived results but also new studies, including multi--robot exploration studies, and derive conclusions regarding the specific role of each algorithm and the considered future research directions. 

The report structure is as follows. In Section~\ref{sec:sec2} each of the exploration and inspection methods is overviewed. A discussion on the role of each method takes place in Section~\ref{sec:sec4}, followed by the description of an architecture for their combination. Finally conclusions and remarks on future work takes place in Section~\ref{sec:sec5}.

\section{EXPLORATION \& INSPECTION STRATEGIES}\label{sec:sec2}

A set of algorithms addressing the problems of inspection and exploration were proposed and are summarized below. 

\subsection{Optimal Inspection Planning}

\begin{figure}[h!]
\centering
  \includegraphics[width=0.99\columnwidth]{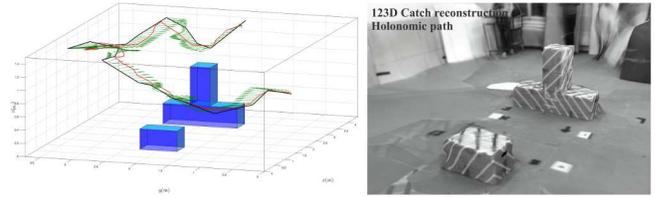}
\caption{Indicative inspection path planning experimental result using the RRTOT method for the case of a hexacopter aerial robot considered to be flying holonomic paths. The path length is $9.1\textrm{m}$ and it was computed after several minutes of operation of the RRTOT algorithm. Video: \texttt{https://youtu.be/e7ljyDM9h8o} }
\label{fig:rrtot_res}
\end{figure}

In the literature, many contributions have been made towards addressing the challenges of coverage planning. Within the most recent contributions, those that employ a two--step optimization scheme proved to be more versatile with respect to the inspection scenario. In a first step, such algorithms compute the minimal set of viewpoints that cover the whole structure which corresponds to solving an Art Gallery Problem (AGP). As a second step, the shortest connecting tour over all these viewpoints has to be computed, which is the Traveling Salesman Problem (TSP). However, the general approach of breaking the problem of finding full coverage paths into that of finding a minimal set of viewpoints and only afterwards perform tour optimization does not --in general-- guarantee inspection path optimality. Furthermore, in cases of nonholonomic vehicles and presence of obstacles, such methods can also lead to overall infreasible solutions. To overcome these limitations, a new algorithm called Rapidly--exploring Random Tree Of Trees (RRTOT) was proposed and employs sampling--based methods and a meta--tree structure consisting of multiple RRT$^{\star}$--trees to find admissible paths with decreasing cost. Using this approach, RRTOT does not suffer from the limitations of strategies that seperate the problem into that of finding the solution to the AGP and afterwards solving the derived TSP. Essentially, RRTOT relies on a sampling--based algorithm that generates trees that root from previously random sampled vertices of other trees. In that sense, this tree--of--trees structure (which is essentially a connected forest) has the capacity to arbitrarily vary coverage path topologies and from that ensures that the optimal solution can be found. In our relevant paper~\cite{bircher_robotica}, the principles of incremental solution derivation and asymptotic optimality are proven. The algorithm supports both holonomic as well as nonholonomic systems, and also further accounts for sensor model constraints. Figure~\ref{fig:rrtot_res} presents one of the relevant experimental results for the case of a multirotor aerial robot. The recorded flight can be found at~\url{https://youtu.be/e7ljyDM9h8o}.

%

\subsection{Efficient Structural Inspection Planning}

\begin{figure}[h!]
\centering
  \includegraphics[width=0.9\columnwidth]{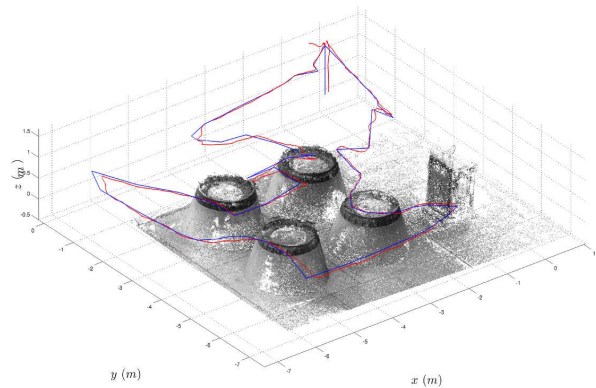}
\caption{Experimental study of the inspection of a subset of the ETH Polyterrasse truncated cones using SIP. The inspection path was computed based on a rough CAD model and the polyhedric obstacle was also included. The path cost is $167.3\textrm{s}$ for maximum forward velocity of $0.25\textrm{m/s}$ and maximum yaw rate equal to $0.5\textrm{rad/s}$. Video: \texttt{https://youtu.be/5kI5ppGTcIQ}}
\label{fig:sip_res}
\end{figure}

Aiming towards \textit{efficient} derivation of full coverage, admissible, optimized although not necessarily true optimal inspection path planning, our team further proposed the Structural Inspection Planner (SIP)~\cite{SIP_AURO_2015,BABOOMS_ICRA_15}, an algorithm that retains a two--step optimization paradigm but contrary to trying to find a minimal set of viewpoints in the AGP, it rather tries to sample them such that the connecting path is short while ensuring coverage. This is driven by the idea that with a continuously sensing sensor, the number of viewpoints (and if this is minimal or not) is not necessarily important but mostly their configuration in space, which has to be such that short and full coverage paths are provided. To achieve its goal, SIP iterates between a step that samples a new set of viewpoint configurations and a second step within which it computes collision--free paths and performs tour optimization. One viewpoint is sampled for each subset of the structure (e.g. for each face of a mesh representation of the structure) and a convex optimization technique ensures the visibility of the associated subset of the overall manifold. As the algorithm iteratively executes these two steps, it manages to find improved solutions - a goal that is further assisted by a set of heuristics~\cite{SIP_AURO_2015}. Both holonomic and nonholonomic vehicles are supported, the constraints of the sensor model are respected, while implementations for mesh--based and octomap--based representations of the structure are available. The overall implementation is open--sourced~\cite{kOpt_CodeRelease} and available as a Robot Operating System (ROS) package. Figure~\ref{fig:sip_res} presents an indicative inspection and $3\textrm{D}$ reconstruction result. The recorded flight video can be found at~\url{https://youtu.be/5kI5ppGTcIQ}.

%
%

\subsection{Uniform Coverage Inspecetion Planning}

\begin{figure}[h!]
\centering
  \includegraphics[width=0.99\columnwidth]{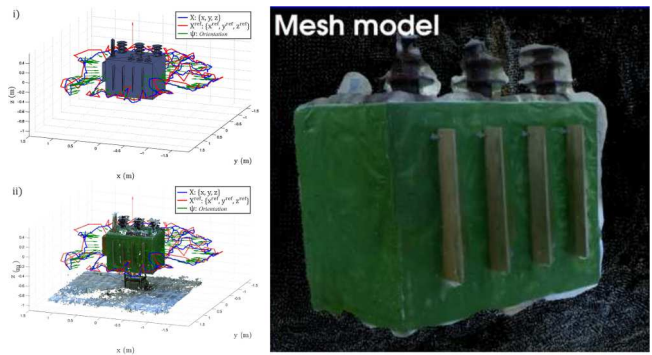}
\caption{Indicative uniform coverage inspection path planning experimental result using the UC$3$D method using a quadrotor aerial robot that relies on a monocular camera/RGB--D localization and mapping pipeline. The power transformed mesh has been subsampled to $134$ faces and $69$ vertices, while the camera mounting is considered to be with 15 degrees pitch down and the minimum inspection distance is set to $0.35\textrm{m}$. Video: \texttt{https://youtu.be/Gg9qsF3y8IU} }
\label{fig:uc3d_res}
\end{figure}

In real--life inspection operations, uniform coverage with equal focus on the details is one of the often desired properties. To aproach this problem, an algorithm that exploits the uniformity properties of Voronoi--based meshing techniques was proposed. The specific method of ``uniform coverage $3\textrm{D}$ structural inspection path planning'' (UC$3$D) iteratively loads lower--fidelity meshes of the structure to be inspected (by subsampling), computes a set of viewpoints with each one of them ensuring the inspection of one of the faces from a similar distance and perceiving angle and finds the optimal tour among them. Viewpoint derivation is achieved by randomly sampling within the subset of the configuration space that allows ``uniform'' inspection, while path feasibility is supported by verifying connectivity with the neighboring viewpoints subject to any nonholonomic constraints. As long as the algorithm cannot find an overall feasible solution at one of its iteration, the process is repeated until randomization leads to solution feasibility. Figure~\ref{fig:uc3d_res} presents an indicative experimental result, while the recorded flight video is available at \url{https://youtu.be/Gg9qsF3y8IU}.

%
%

\subsection{Autonomous Exploration and Localizability}

\begin{figure}[h!]
\centering
  \includegraphics[width=0.99\columnwidth]{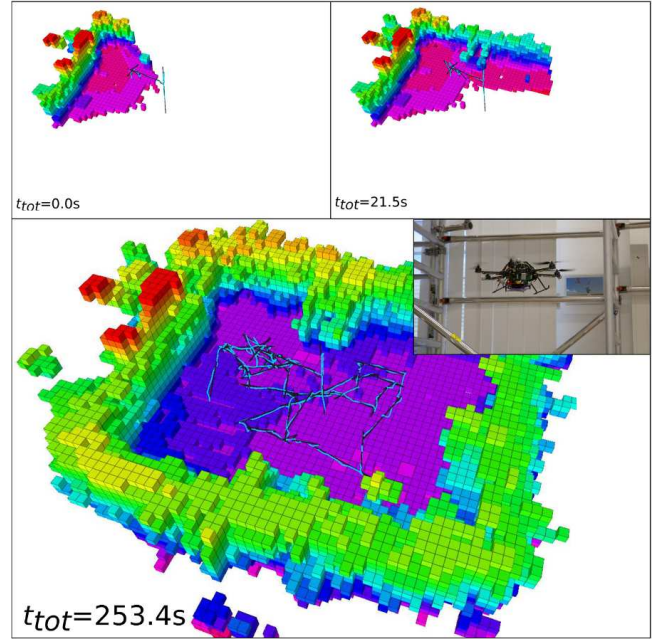}
\caption{Exploration experiment in a closed room. The colored voxels (color selected based on height) represent occupied parts of the occupancy map. The computed path is shown with black color, while the experimentally recorded path of the robot is shown with light blue. Video: \texttt{https://youtu.be/D6uVejyMea4} }
\label{fig:nbvp_res}
\end{figure}

Autonomous exploration planning refers to the capacity of a robot to map a previously unknown environment.  Early work includes \cite{connolly1985determination}, where good ``next--best--views'' are determined in order to cover a given structure. Advanced versions were recently presented~\cite{vasquez2014volumetric}, while the method of frontiers--based planning corresponds to one of the most widely used exploration strategies. Within our work in~\cite{NBVP_ICRA_16,bircher2016receding}, a receding horizon approach to the problem of Next--Best--Ciew Planning (NBVP) is proposed and experimentally verified. The views are sampled as nodes in a random tree, the edges of which directly give a path to follow such that the viewpoints are sequentially reached. At every step, a finite--depth tree of views is sampled but only the first step is executed by the robot, while the whole process is repeated at the next iteration. This receding horizon strategy improves and robustifies the exploratory behavior of the robot. Figure~\ref{fig:nbvp_res} presents an indicative experimental result. It is noted that this planner is also open--sourced~\cite{nbvpCodeRelease} and accompanied by an open dataset~\cite{nbvp_DatasetRelease}. A relevant experiment is recorded and is available at~\url{https://youtu.be/D6uVejyMea4}.

%
%

\begin{figure}[h!]
\centering
  \includegraphics[width=0.99\columnwidth]{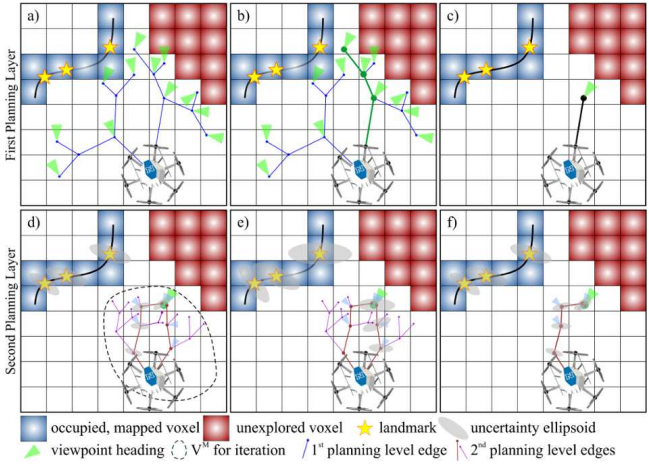}
\caption{2D representation of the two--steps uncertainty--aware exploration and mapping planner. The first planning layer samples the path with the maximum exploration gain. The viewpoint configuration of the first vertex of this path becomes the reference to the second planning layer. Then this step, samples admissible paths that arrive to this configuration, performs belief propagation along the tree edges, and selects the one that provides minimum uncertainty over the robot pose and tracked landmarks. The video in \texttt{https://youtu.be/iveNtQyUut4} presents the overall experiment.}
\label{fig:rhem_steps}
\end{figure}

\begin{figure}[h!]
\centering
  \includegraphics[width=0.99\columnwidth]{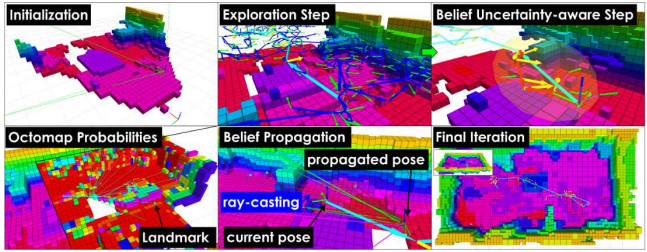}
\caption{nstances of an exploration and mapping experiment in a closed room with a challenging geometry. The initial phase of the exploration is dominated by yawing motions. Especially when long paths are selected, the second planning layer identifies alternative paths that optimize the robot belief. Furthermore, as shown the probabilistic backend of octomap is maintained to allow the computation of the $\mathbf{ReobservationGain}$, while during belief propagation, visibility check for the tracked landmarks takes place. The result is a consistent $3\textrm{D}$ map despite the size and the challenges of the environment.}
\label{fig:arena_experiment}
\end{figure}

Furthermore, in our work in~\cite{RHEM_ICRA_2017}, the problems of autonomous exploration and robot localizability are addressed together. In particular, a localization uncertainty--aware Receding Horizon Exploration and Mapping (RHEM) planner is proposed. The RHEM planner relies on a two--step, receding horizon, belief space--based approach. At first, in an online computed random tree, the algorithm identifies the branch that optimizes the amount of new space expected to be explored. The first viewpoint configuration of this branch is selected, but the path towards it is decided through a second planning step. Within that, a new tree is sampled, admissible branches arriving at the reference viewpoint are found and the robot belief about its state and the tracked landmarks is propagated. As system state the concatenation of the robot states and tracked landmarks (visual features) is considered. Then, the branch that minimizes the localization uncertainty, as factorized using the D--optimality (D--opt) of the pose and landmarks covariance is selected. The corresponding path is conducted by the robot and the process is iteratively repeated. Figure~\ref{fig:rhem_steps} illustrates the basic steps of this planner. Figure~\ref{fig:arena_experiment} presents indicative experimental results and the video in \url{https://youtu.be/iveNtQyUut4} demonstrates the overall experiment. 

%
%

%
%

\subsection{Contact--based Inspection}

\begin{figure}[h!]
\centering
  \includegraphics[width=0.80\columnwidth]{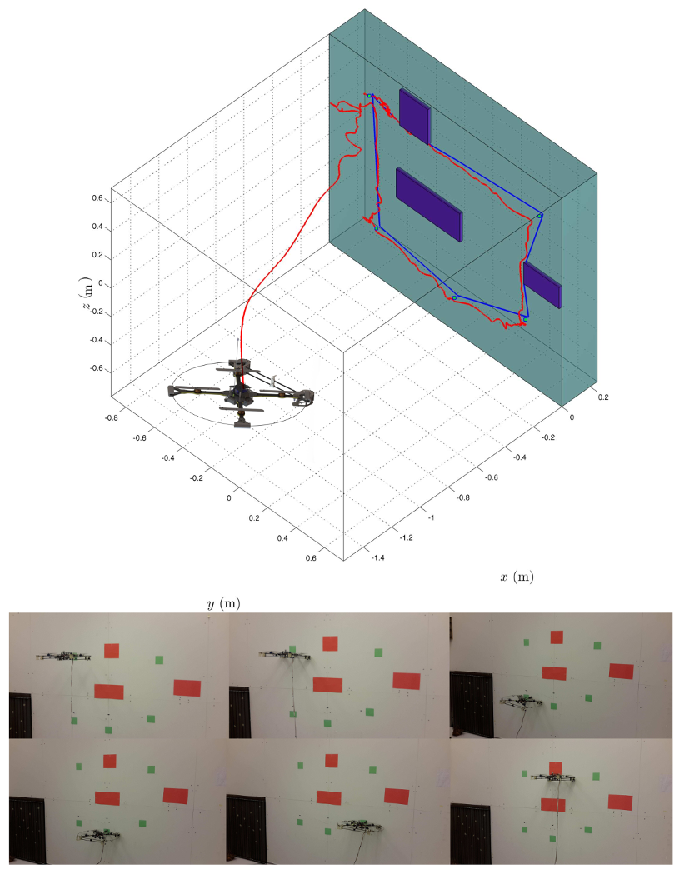}
\caption{Contact--based inspection mission using the CIPC strategy. The robot has to visit the specified points of interest while avoiding any obstacles of the environment. As shown ``obstacle'' areas have been attached on the wall. The CIPC strategy successfully establishes contact and subsequently executes the optimized in--contact inspection path. Video: \texttt{https://youtu.be/lDpHNEB66wE} }
\label{fig:cip_res}
\end{figure}

Given that a structure is inspected and its $3\textrm{D}$ reconstruction has been succesfully derived, a next possible step within an infrastructure monitoring application may require contact--based inspection to conduct non--destructive testing for structural integrity aspects such as gas pipe wall thickness or measurements. These processes are conducted using sensors such as ultrasound probes which require physical contact with the structure. This fact, motivated the research efforts of our team and of the commmunity to address the problem of flight control during physical interaction. In response to this need, we developed a Hybrid Model Predictive Control (HMPC) approach that relies on a hybrid model of the aerial robot dynamics using essentially different dynamic modes during the free--flight and physical interaction phases of the operation~\cite{DABS_ICRA_14,AHS_ICRA_13}. The HMPC approach ensures stability during the mode switching, robust execution of physical interaction tasks and high--performance free--flight. Building on top of this capacity, a framework that allows the user to select a set of points to be inspected on the physical surface and then finding the optimal route among them was proposed~\cite{ADBS_AURO_2015} and the overall method is called Contact--based Inspection Planning and Control (CIPC). This framework also allows to overcome an obstacle on the physical surface also by undocking from it and re--docking at the next point of interest. Figure~\ref{fig:cip_res} presents an indicative result using a multicopter aerial robot, while the recorded flight video is available at \url{https://youtu.be/lDpHNEB66wE}. 

%
%

\section{A UNIFYING ARCHITECTURE}\label{sec:sec4}

\begin{figure}[h!]
\centering
  \includegraphics[width=0.99\columnwidth]{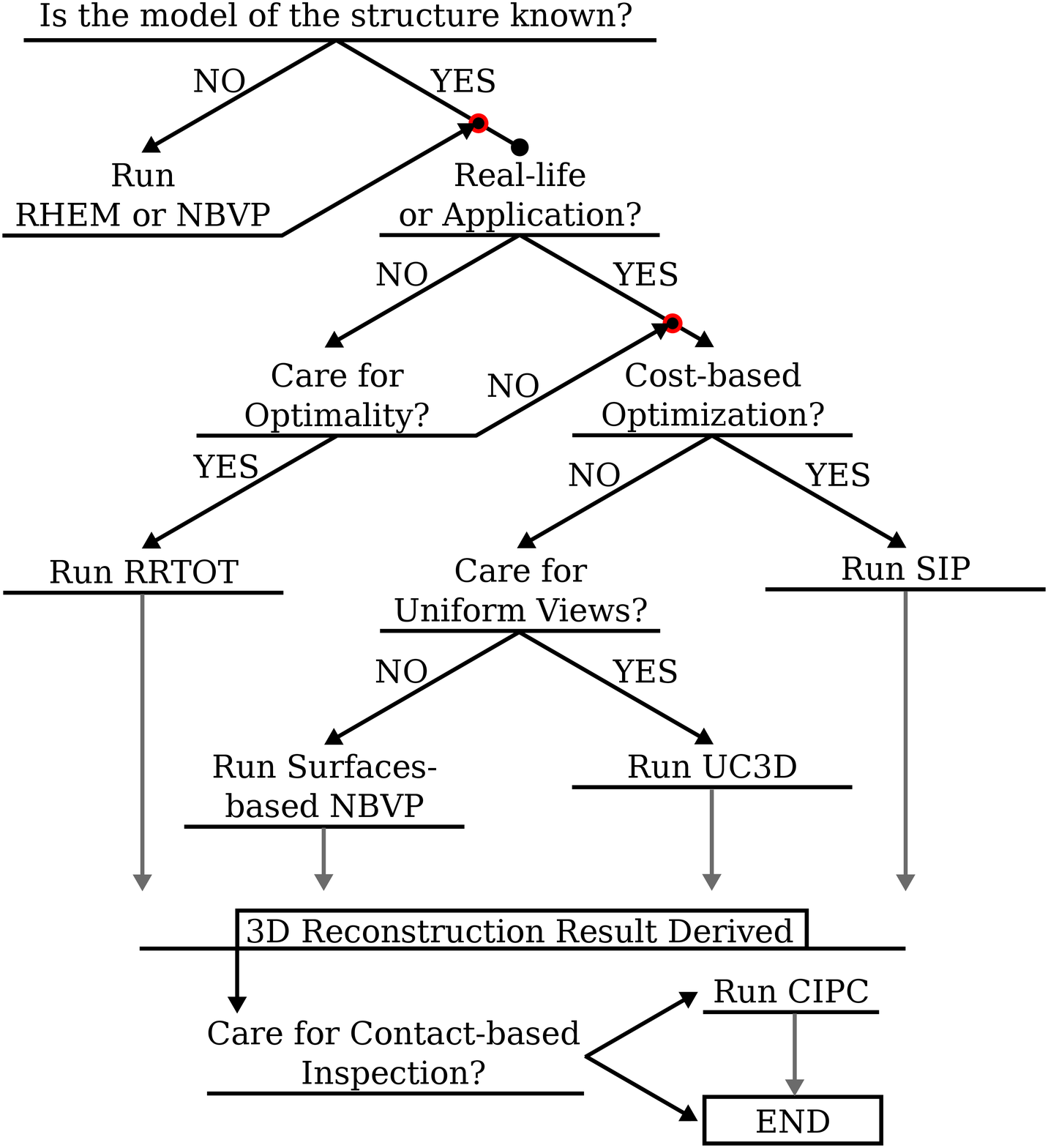}
\caption{Proposed architecture for the combination of the different structural inspection and exploration algorithms of the proposed planning ensemble towards a complete solution that addresses the key end--user requirements regarding the operation in known or unknown environments, as well as the different operation requirements regarding the data to be used for the $3\textrm{D}$ reconstruction process. }
\label{fig:planning_ensemble}
\end{figure}

A planning ensemble for inspection and exploration has been proposed and experimentally verified towards realizing the aerial robotic worker for inspection operations. Among the three algorithms for inspection, RRTOT is characterized by optimality but very expensive computations, UC$3$D focuses primarily on uniform coverage, while SIP provides a rather balanced solution characterized by optimized cost and limited computational cost and needs. All these three algorithms require that a geometrical model of the structure to be inspected is known a priori. On the contrary, NBVP and RHEM assume no prior knowledge of the environment and enable its autonomous exploration. While RRTOT, SIP and UC$3$D are global planners, NBVP and RHEM are local planning solutions that reactively compute the next--best--viewpoint of the robot given its online computed $3\textrm{D}$ reconstruction of the previously unknown environment. The RHEM planner goes further to identify the trajectory that visits the best exploration viewpoint while maintaining low localization uncertainty. Given the different role and features of these algorithms, they can correspond to a relatively complete, real--life, structural inspection solution through their combination. Figure~\ref{fig:planning_ensemble} presents the proposed architecture for the utilization of the aerial robotic workers inspection planning ensemble.

%
%

\section{OPEN SOURCE CONTRIBUTIONS}\label{sec:sec4}

To accelerate the utilization of autonomous exploration and inspection technologies, support the community developments and overall lead to living contributions, a subset of these algorithms have been opens-sourced. This refers to the SIP planner~\cite{kOpt_CodeRelease}, NBVP~\cite{nbvpCodeRelease} and RHEM~\cite{rhemCodeRelease}. All repositories are also accompanied by experimental datasets. Further relevant data can be found in~\cite{Oettershagen_FSR2015}. 

\section{CONCLUSIONS \& FUTURE WORK}\label{sec:sec5}

A planning ensemble that enables the realization of the autonomous aerial robotic worker for inspection operations is presented. The set of algorithms contains solutions for the optimized inspection given a prior geometric model of the structure, as well as fully autonomous solutions that are futher localization belief uncertainty--aware. A summary of the functioning principle of each algorithm is presented, in combination with characteristic results and discussion on its main properties. Finally, an architecture of their combination is presented in order to solve real--life infrastructure inspection challenges for which a previous model might --or might not-- be available. 



\begin{thebibliography}{10}
\providecommand{\url}[1]{#1}
\csname url@rmstyle\endcsname
\providecommand{\newblock}{\relax}
\providecommand{\bibinfo}[2]{#2}
\providecommand\BIBentrySTDinterwordspacing{\spaceskip=0pt\relax}
\providecommand\BIBentryALTinterwordstretchfactor{4}
\providecommand\BIBentryALTinterwordspacing{\spaceskip=\fontdimen2\font plus
\BIBentryALTinterwordstretchfactor\fontdimen3\font minus
  \fontdimen4\font\relax}
\providecommand\BIBforeignlanguage[2]{{%
\expandafter\ifx\csname l@#1\endcsname\relax
\typeout{** WARNING: IEEEtran.bst: No hyphenation pattern has been}%
\typeout{** loaded for the language `#1'. Using the pattern for}%
\typeout{** the default language instead.}%
\else
\language=\csname l@#1\endcsname
\fi
#2}}

\bibitem{nsf_asce_card2013}
\BIBentryALTinterwordspacing
{American Society of Civil Engineers}, ``{ASCE 2013 Report Card on America's
  Infrastructure}.'' [Online]. Available:
  \url{http://www.infrastructurereportcard.org/}
\BIBentrySTDinterwordspacing

\bibitem{bircher_robotica}
{A. Bircher, K. Alexis, U. Schwesinger, S. Omari, M. Burri, and R. Siegwart},
  ``An incremental sampling-based approach to inspection planning: The
  rapidly-exploring random tree of trees,'' 2015.

\bibitem{SIP_AURO_2015}
{A. Bircher, M. Kamel, K. Alexis, M. Burri, P. Oettershagen, S. Omari, T.
  Mantel and R. Siegwart}, ``\BIBforeignlanguage{English}{Three-dimensional
  coverage path planning via viewpoint resampling and tour optimization for
  aerial robots},'' \emph{\BIBforeignlanguage{English}{Autonomous Robots}}, pp.
  1--25, 2015.

\bibitem{BABOOMS_ICRA_15}
\BIBentryALTinterwordspacing
{A. Bircher, K. Alexis, M. Burri, P. Oettershagen, S. Omari, T. Mantel and R.
  Siegwart}, ``Structural inspection path planning via iterative viewpoint
  resampling with application to aerial robotics,'' in \emph{IEEE International
  Conference on Robotics and Automation (ICRA)}, May 2015, pp. 6423--6430.
  [Online]. Available:
  \url{https://github.com/ethz-asl/StructuralInspectionPlanner}
\BIBentrySTDinterwordspacing

\bibitem{kOpt_CodeRelease}
\BIBentryALTinterwordspacing
{A. Bircher, and K. Alexis}, ``{Structural Inspection Planner Code Release}.''
  [Online]. Available:
  \url{https://github.com/ethz-asl/StructuralInspectionPlanner}
\BIBentrySTDinterwordspacing

\bibitem{connolly1985determination}
C.~Connolly \emph{et~al.}, ``The determination of next best views,'' in
  \emph{Robotics and Automation. Proceedings. 1985 IEEE International
  Conference on}, vol.~2.\hskip 1em plus 0.5em minus 0.4em\relax IEEE, 1985,
  pp. 432--435.

\bibitem{vasquez2014volumetric}
J.~I. Vasquez-Gomez, L.~E. Sucar, R.~Murrieta-Cid, and E.~Lopez-Damian,
  ``Volumetric next best view planning for 3d object reconstruction with
  positioning error,'' \emph{Int J Adv Robot Syst}, vol.~11, p. 159, 2014.

\bibitem{NBVP_ICRA_16}
\BIBentryALTinterwordspacing
{A. Bircher, M. Kamel, K. Alexis, H. Oleynikova and R. Siegwart}, ``Receding
  horizon "next-best-view" planner for 3d exploration,'' in \emph{IEEE
  International Conference on Robotics and Automation (ICRA)}, May 2016.
  [Online]. Available: \url{https://github.com/ethz-asl/nbvplanner}
\BIBentrySTDinterwordspacing

\bibitem{bircher2016receding}
A.~Bircher, M.~Kamel, K.~Alexis, H.~Oleynikova, and R.~Siegwart, ``Receding
  horizon path planning for 3d exploration and surface inspection,''
  \emph{Autonomous Robots}, pp. 1--16, 2016.

\bibitem{nbvpCodeRelease}
\BIBentryALTinterwordspacing
{A. Bircher, and K. Alexis}, ``{Receding Horizon Next Best View Planner}.''
  [Online]. Available: \url{https://github.com/ethz-asl/nbvplanner}
\BIBentrySTDinterwordspacing

\bibitem{nbvp_DatasetRelease}
\BIBentryALTinterwordspacing
{A. Bircher, M. Kamel, K. Alexis, H. Oleynikova and R. Siegwart}, ``{Receding
  Horizon Next Best View Planner Dataset}.'' [Online]. Available:
  \url{https://github.com/ethz-asl/nbvplanner/wiki/Example-Results}
\BIBentrySTDinterwordspacing

\bibitem{RHEM_ICRA_2017}
\BIBentryALTinterwordspacing
{C. Papachristos, S. Khattak, and K. Alexis}, ``Uncertainty--aware receding
  horizon exploration and mapping using aerial robots,'' in \emph{IEEE
  International Conference on Robotics and Automation (ICRA)}, May 2017.
  [Online]. Available: \url{https://github.com/unr-arl/rhem_planner}
\BIBentrySTDinterwordspacing

\bibitem{DABS_ICRA_14}
G.~Darivianakis, K.~Alexis, M.~Burri, and R.~Siegwart, ``Hybrid predictive
  control for aerial robotic physical interaction towards inspection
  operations,'' in \emph{Robotics and Automation (ICRA), 2014 IEEE
  International Conference on}, May 2014, pp. 53--58.

\bibitem{AHS_ICRA_13}
K.~Alexis, C.~Huerzeler, and R.~Siegwart, ``Hybrid modeling and control of a
  coaxial unmanned rotorcraft interacting with its environment through
  contact,'' in \emph{2013 International Conference on Robotics and Automation
  (ICRA)}, Karlsruhe, Germany, 2013, pp. 5397--5404.

\bibitem{ADBS_AURO_2015}
\BIBentryALTinterwordspacing
K.~Alexis, G.~Darivianakis, M.~Burri, and R.~Siegwart,
  ``\BIBforeignlanguage{English}{Aerial robotic contact-based inspection:
  planning and control},'' \emph{\BIBforeignlanguage{English}{Autonomous
  Robots}}, pp. 1--25, 2015. [Online]. Available:
  \url{http://dx.doi.org/10.1007/s10514-015-9485-5}
\BIBentrySTDinterwordspacing

\bibitem{rhemCodeRelease}
\BIBentryALTinterwordspacing
{C. Papachristos, S. Khattak, and K. Alexis}, ``{Uncertainty--aware Receding
  Horizon Exploration and Mapping Planner}.'' [Online]. Available:
  \url{https://github.com/unr-arl/rhem_planner}
\BIBentrySTDinterwordspacing

\bibitem{Oettershagen_FSR2015}
{P. Oettershagen, T. Stastny, T. Mantel, A. Melzer, K. Rudin, G. Agamennoni, K.
  Alexis, and R. Siegwart}, ``Long-endurance sensing and mapping using a
  hand-launchable solar-powered uav,'' June 2015.

\end{thebibliography}

\end{document}